\documentclass{article}
\usepackage{CJKutf8}
\usepackage{amsmath}
\usepackage{txfonts}
\usepackage{amsfonts}

\usepackage{pdfpages}
\usepackage{svg} 
\usepackage{PRIMEarxiv}
\usepackage{indentfirst} 
\usepackage[utf8]{inputenc} 
\usepackage[T1]{fontenc}    
\usepackage{hyperref}       
\usepackage{url}            
\usepackage{booktabs}       
\usepackage{amsfonts}       
\usepackage{nicefrac}       
\usepackage{microtype}      
\usepackage{lipsum}
\usepackage{fancyhdr}       
\usepackage{geometry}
\usepackage{booktabs}
\usepackage{makecell} 

\graphicspath{{media/}}     

\title{Fused attention mechanism-based ore sorting network
\thanks{\textit{\underline{Citation}}: 
\textbf{}} 
}

\author{
  Junjiang Zhen,  \\
  Department of Mathematical and Information Sciences\\
  Hebei University \\
  baoding,China\\
  \texttt{20227014020@stumail.hbu.edu.cn} \\
   \And
  Bojun Xie \\
  Department of Mathematical and Information Sciences \\
  Hebei University \\
  baoding,China\\
  \texttt{xiebojun@hbu.edu.cn} \\
}

\begin{document}
\maketitle

\begin{abstract}
Deep learning has had a significant impact on the identification and classification of mineral resources, especially playing a key role in efficiently and accurately identifying different minerals, which is important for improving the efficiency and accuracy of mining. However, traditional ore sorting methods often suffer from inefficiency and lack of accuracy, especially in complex mineral environments. To address these challenges, this study proposes a method called OreYOLO, which incorporates an attentional mechanism and a multi-scale feature fusion strategy, based on ore data from gold and sulfide ores. By introducing the progressive feature pyramid structure into YOLOv5 and embedding the attention mechanism in the feature extraction module, the detection performance and accuracy of the model are greatly improved. In order to adapt to the diverse ore sorting scenarios and the deployment requirements of edge devices, the network structure is designed to be lightweight, which achieves a low number of parameters (3.458M) and computational complexity (6.3GFLOPs) while maintaining high accuracy (99.3\% and 99.2\%, respectively). In the experimental part, a target detection dataset containing 6000 images of gold and sulfuric iron ore is constructed for gold and sulfuric iron ore classification training, and several sets of comparison experiments are set up, including the YOLO series, EfficientDet, Faster-RCNN, and CenterNet, etc., and the experiments prove that OreYOLO outperforms the commonly used high-performance object detection of these architectures.
\end{abstract}


\section{Introduction}
\setlength{\parindent}{2em} 
In the current global context of structural changes in mineral resources, and facing the challenge of depleting high-grade ore reserves, the mining industry is inevitably shifting towards the development of more complex and lower-grade ore resources. This transition not only imposes stricter requirements on processing technologies but also presents new challenges in environmental protection and economic cost management. Driven by this trend, the development of efficient and sustainable ore sorting technologies has become an urgent task in the mining industry.

The rapid development of deep learning technology has brought new possibilities for ore classification and identification, especially through image classification methods based on Convolutional Neural Networks (CNNs). Due to their superior performance in analyzing and processing color texture features, CNNs have demonstrated their potential in automated and efficient ore identification tasks. Su et al. first applied LeNet-5 to a binary classification task of coal and gangue, training it with 20,000 images to achieve an accuracy of 95.88\% \cite{su2018research}. Wang et al. utilized the Wu-VGG19 transfer network for binary classification of scheelite ore and surrounding rocks, addressing the problem of ineffective recognition of scheelite ore and waste rock in scheelite ore image recognition methods \cite{liguan2020beneficiation}. Using the Inception-v3 network, Zhang et al. established a transfer learning model for mineral micrograph images, achieving an accuracy of 90.9\% in quartz and feldspar micrograph images \cite{zhang2019intelligent}. Liu et al. employed ResNet-50 to classify granite, diorite, and gabbro, achieving excellent results with an accuracy of 95 \cite{liu2021deep}. These methods mostly focus on classification and subsequent localization tasks.

With the evolution of technology, the design, deployment, and application of high-performance visual models have become feasible even in complex working conditions. Liu et al. proposed a novel lightweight network called LOSN (Lightweight Ore Sorting Network), specifically designed for the classification and localization of bituminous coal and anthracite. This method is integrated with a robotic arm, achieving an end-to-end solution for mineral sorting. The research team conducted detailed analysis of the LOSN model in terms of model depth, architecture, and accuracy, providing comprehensive reference for the application of deep learning in coal classification \cite{liu2023losn}. Zhao et al. introduced the K-means clustering algorithm and integrated the SE attention mechanism into YOLOv4, achieving a remarkable average accuracy of 99\% in bauxite ore sorting tasks after 7000 training iterations, while maintaining inference speed faster than 0.05 seconds. This achievement not only significantly improves the speed and accuracy of bauxite ore sorting, but also greatly enhances the efficiency and precision of the bauxite ore industry \cite{zhao2022machine}. Zhang et al. proposed a brightness equalization algorithm to reduce the impact of illumination intensity on recognition accuracy, and used YOLOv5 to identify samples, ultimately achieving a 95.6\% accuracy rate on 50 common minerals \cite{zhang2022mineral}. Huang et al. utilized the YOLO-v3 model as a base detector, initialized with pretrained weights on the MS COCO dataset. Subsequently, they constructed an AL framework model embedded with YOLO-v3 for autonomous detection of bauxite slurry \cite{huang2020weakly}. Xiao et al. integrated the CECA attention module into YOLOv4 for foreign object detection in gold ore veins, achieving 90.7\% accuracy \cite{xiao2023mining}.

Closely following the strides of computing and imaging technologies, intelligent ore sorting techniques have been widely deployed in the industrial sector. Leveraging advanced optical systems to capture images and extract feature information, these systems achieve precise classification under diverse environmental conditions. They stand out for their advantages of low cost, high efficiency, absence of radiation hazards, and ease of installation \cite{luo2022review}. Currently, these technologies primarily revolve around machine learning and deep learning image processing techniques, such as decision trees \cite{breiman1996bagging}\cite{breiman2017classification}, naive Bayes \cite{wang2007naive}, K-nearest neighbors \cite{liu2019enhanced}\cite{cover1967nearest}, and support vector machines (SVM) \cite{cortes1995support}, which have found extensive applications in the field of ore image classification. Nevertheless, machine learning-based methods often rely on high-definition images and may lack stability in maintaining image quality in dynamic environments. Additionally, their dependency on domain expertise for feature extraction increases the practical application threshold, thereby limiting their scope and deployment speed.

Therefore, this paper proposes a novel method for sorting gold and pyrite ores by offline ore data collection and training using YOLOv5 network. Considering the real-time requirements of the actual operation, the network is optimized for lightweighting in the training process. In addition, to solve the problem of insufficient feature extraction in the lightweight network, we embed the scale attention (EMA) mechanism into the backbone network of YOLOv5, which enables the network to extract richer features. Meanwhile, an efficient asymptotic feature pyramid network (AFPN) structure is used to avoid losing or degrading information during the feature fusion process and to retain more effective features for fusion. In the experimental part, this study compares OreYOLO with ten common deep learning-based target detection networks (e.g., YOLO series, EfficientDet series, Faster-RCNN, and CenterNet) in terms of accuracy, model size, computational complexity, and inference speed. The experimental results show that OreYOLO performs well on the gold and pyrite datasets with an accuracy of 86.6\%, which is 3.2\% higher than the second place Faster-RCNN (83.4\%). In terms of computational complexity, model size, and inference speed, OreYOLO has a relatively small number of parameters, lower GFLOPs, and achieves higher FPS of 3.458M parameters, 6.3G GFLOPs, and 79.07FPS, respectively, which outperforms most target detection networks.

\section{ Details of YOLOv5 Network}
As one of the most cutting-edge object detection algorithms currently, YOLOv5 is highly praised for its outstanding speed and accuracy performance. Its core concept revolves around dividing the entire image into grids, allowing each grid cell to predict the category and spatial location of objects it contains. Subsequently, target boxes are filtered based on the Intersection over Union (IoU) between predicted boxes and ground truth boxes, ultimately outputting the category and location information for each target. As illustrated in Figure 1, the network architecture of YOLOv5 can be divided into four key components: the input layer, the backbone network, the neck network, and the prediction layer.

At the input stage, the model's input characteristics are optimized through Mosaic data augmentation techniques and adaptive anchor box calculation. In the backbone network section, the Cross Stage Partial Bottleneck (CSP) structure is employed \cite{wang2020cspnet}. This structure divides the input feature map into two paths: one path for direct transmission, while the other path undergoes a series of convolution operations. Finally, the two paths are merged, reducing computational costs and parameter count while ensuring efficient information flow and feature extraction capability.
\begin{figure}[htb]
    \centering
     \includegraphics[width=1\textwidth, trim=6.5cm 7.7cm 6.5cm 8.2cm, clip]{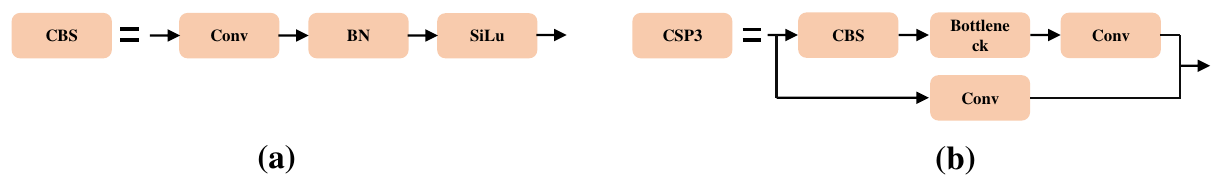}
    \caption{(a) The CBS block consists of a convolutional layer, a normalization layer and a SiLu activation function. (b) The CSP3 block consists of a CBS block with residual paths and a Bottleneck structure}
    \label{fig:(a) CBS块由一个卷积层，一个归一化层和一个SiLu激活函数组成.(b)CSP3块由一个带残差路径的CBS块和一个Bottleneck结构组成}
\end{figure}
In the neck section, a combination of Feature Pyramid Network (FPN) \cite{lin2017feature} and Path Aggregation Network (PAN) \cite{liu2018path} is applied, as illustrated in Figure 2. The FPN layer passes rich semantic information along the top-down path, achieving the transmission and fusion of high-level feature information through upsampling operations, thereby providing enhanced feature maps for the prediction process. Meanwhile, the PAN layer passes precise localization features along the bottom-up path, facilitating the classification and localization accuracy of the detection task.

The prediction layer adopts a multi-level feature fusion strategy. This strategy first reduces the channel number and scales down the size of the feature maps output by the backbone network through convolutional operations. Then, it integrates feature maps from different levels to extract richer feature information, thereby significantly improving the model's detection performance.
\begin{figure}[htb]
    \centering
    \includegraphics[width=1\textwidth, trim=2cm 6cm 1cm 4cm, clip]{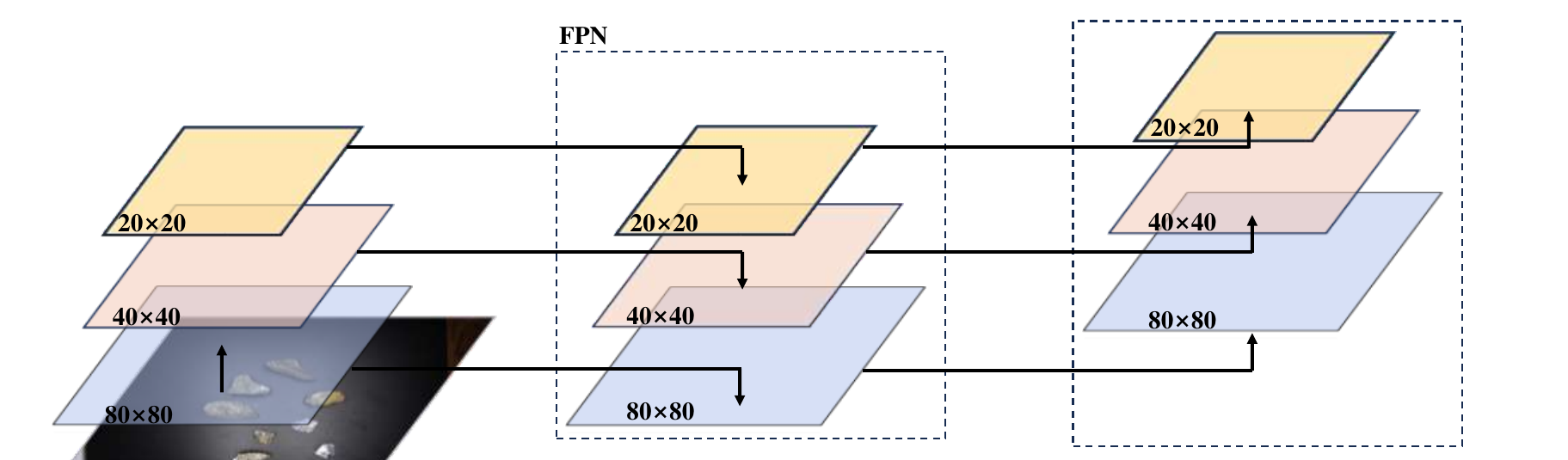}
    \caption{The structure of the Path Aggregation Network (PAN), consisting of Bottom-up feature fusion and Top-down feature fusion,where 20×20, 40×40, and 80×80 are feature map sizes}
    \label{fig:其中，20×20，40×40，80×80为特征图大小}
\end{figure}
\begin{figure}[htb]
    \centering
    \includegraphics[width=1\textwidth, trim=1cm 5cm 1cm 4cm, clip]{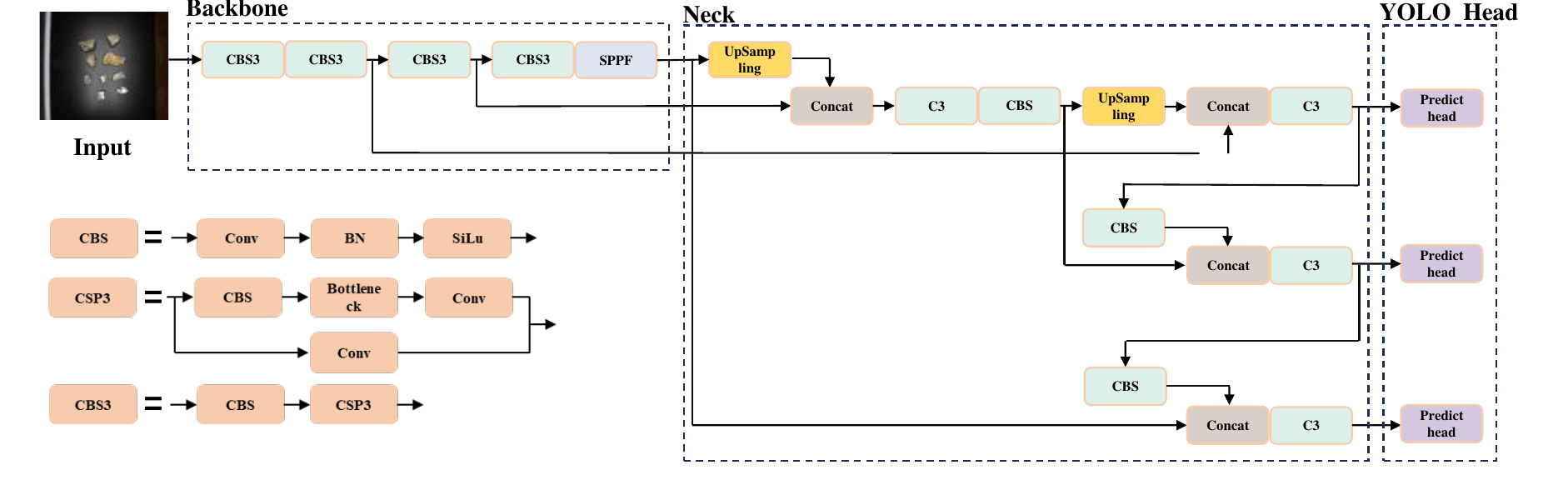}
    \caption{The network structure of YOLOv5 is divided into four main parts, the Input part for image input, the Backbone part for extracting ore features, the Neck part for fusing different layers of features and the YOLO Head part for prediction}
    \label{fig:YOLOv5的网络结构，主要分为四个部分，用于图片输入的Input部分，用于提取矿石特征的BACKbone部分，用于将不同层级特征进行融合的Neck部分和用于预测的YOLO Head部分}
\end{figure}
\section{ Methodology}
\label{sec:headings}
This section will delve into three key aspects of constructing the OreYOLO ore sorting network: implementing attention mechanisms to enhance feature extraction capabilities, integrating progressive feature pyramid structures to enhance feature representation capabilities, and defining OreYOLO's loss function.

\subsection{Attention mechanism}

In this study, an attention mechanism, Efficient Multi-Scale Attention (EMA Attention) \cite{ouyang2023efficient}, is introduced to address the problem of the variability of color texture among different ores, drawing on the human visual physiological perception mechanism. This mechanism is based on the idea of CA Attention \cite{hou2021coordinate}, which adopts a parallel substructure to reduce the sequential processing requirements and depth of the network, while dividing the input into multiple sub-feature groups in the channel dimension to learn rich semantic information, thus effectively improving the generalization ability and recognition accuracy of the model.Efficient Multi-Scale Attention Structure As shown in Fig. 3, it follows the parallel substructure in the CA block to avoid more sequential processing of the network and larger depth and divides the input into multiple sub-features along the channel dimension direction to learn different semantics feature.

\begin{figure}[htb]
    \centering
    \includegraphics[width=1\textwidth, trim=1cm 1cm 4cm 0cm, clip]{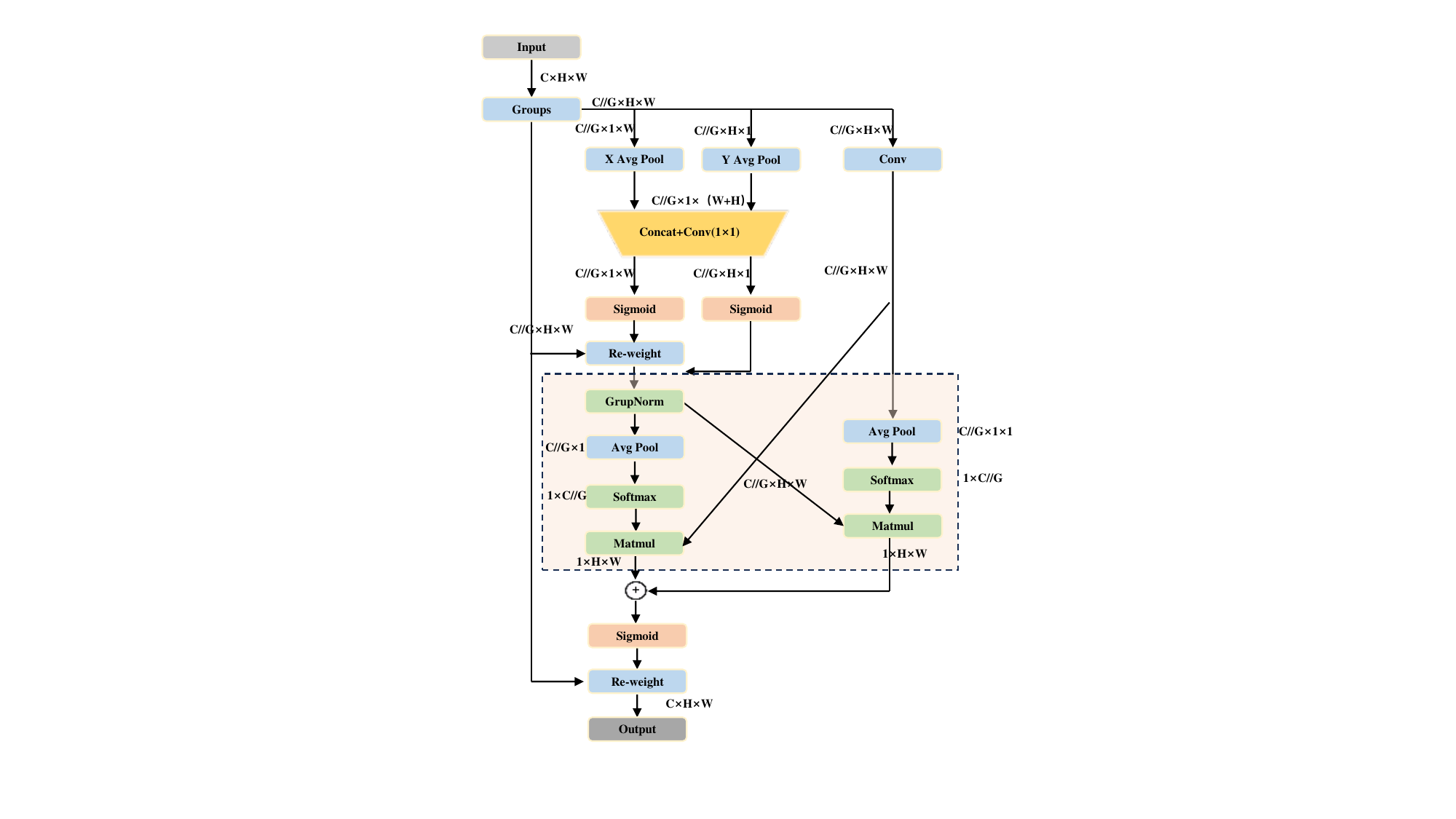}
    \caption{EMA structure diagram, which consists of three parallel paths to extract the attention weight descriptors of the feature map group, where X Avg POOL denotes global pooling in the horizontal direction and Y Avg POOL denotes global pooling in the vertical direction}
    \label{fig:EMA结构图，它由三条并行路径来提取特征图组的注意力权重描述符，其中X Avg POOL表示水平方向的全局池化，Y  Avg POOL 表示垂直方向的全局池化}
\end{figure}
\subsubsection{Feature Grouping} 
For any given input feature map \( X\in \mathbb{R}^{C\times H\times W} \),The EMA mechanism decomposes \(X\) into \(G\) sub-feature groups along the channel dimension 
\(
X=\left[ X_0,X_{\boldsymbol{i}},\ldots,X_{G-1} \right]
\)
,
\(
X_{\boldsymbol{i}}\in \mathbb{R}^{C\varparallel G\times H\times W}
\).
Each group focuses on capturing different semantic levels, and this grouping method enables the model to delve deeper into and utilize the intrinsic semantic features of the input data, further enhancing the expressiveness of the features.
\subsubsection{Feature Grouping} 
EMA adopts three parallel paths to extract attention weight descriptors for feature map groups, optimizing the overall representation of features. Among them, two paths apply 1x1 and 3x3 convolutional kernels respectively, aiming to encode channel information and capture local cross-channel interactions and feature representations at different spatial scales.In addition, EMA encodes channel information by performing 2D global average pooling operations in two spatial directions and aggregates these encoded features. This not only enhances cross-channel interactions but also achieves efficient feature sharing without adding additional dimensions. This strategy adjusts the importance of different channels while accurately preserving spatial structural information, significantly improving the model's recognition accuracy and feature representation capability
\subsubsection{Cross-spatial learning.} 
Cross-spatial learning significantly enhances feature representation capability in computer vision tasks by establishing intrinsic connections between channels and spatial positions. Specifically, EMA utilizes two different convolution branches (1x1 and 3x3) to capture multi-scale features and strengthens inter-channel dependencies through cross-channel interactions. In this process, 2D global average pooling operations on two spatial dimensions encode global spatial information from the output of the 1x1 convolution branch. Then, the global spatial information from the output of the 1x1 branch is encoded, and before the joint activation mechanism of channel features, the output of the minimal branch is directly transformed into the corresponding dimensional shape 
\(
\mathbb{R}_{1}^{\ 1\times C\varparallel G}\times \mathbb{R}_{3}^{\ C\varparallel G\times HW}
\)
the formula for 2D global pooling operation is:
\begin{equation}
\
\text{z}_{c} = \frac{1}{H \times W} \sum_{j}^{H} \sum_{i}^{W} x_{c}(i, j)
\
\end{equation}
Where,
\(
H,W
\)
represent the height and width of the feature map, respectively,
\(x_c\left( i,j \right) \)
is the pixel value at position \(\left( i,j \right)\)on the feature map at channel \(c\)

Through this cross-space learning approach, rich feature information can be aggregated at different scales, generating two spatial attention maps. The first spatial attention map is obtained by encoding global spatial information from the output of the 1x1 convolution branch, while the second spatial attention map is obtained by performing the same process on the output of the 3x3 convolution branch. These two attention maps preserve precise spatial positional information during the aggregation process.Finally, by aggregating these two spatial attention values and applying the Sigmoid function, the model can emphasize pairwise relationships at the pixel level and highlight the global context of all pixels, effectively enhancing the model's attention to higher-level features, thereby improving recognition accuracy and model performance.

\subsection{Progressive Feature Fusion}
\subsubsection{Asymptotic Architecture} 
In object detection tasks, multi-scale features are crucial for encoding objects with scale variations. A common strategy for multi-scale feature extraction is to use classical top-down and bottom-up feature pyramid networks. However, these methods suffer from feature information loss or degradation, impairing the fusion effect between non-adjacent levels.Therefore, OreYOLO introduces the Asymptotic Feature Pyramid Network for Object Detection (AFPN)\cite{yang2023afpn}to support direct interaction between non-adjacent levels.

The AFPN architecture, as shown in Figure 4, specifically, in the top-down feature extraction process, first initiates the fusion process by combining two low-level features with different resolutions, then gradually incorporates high-level features into the fusion process, and finally merges the top-level features of the backbone network, namely the most abstract features.This fusion approach helps to avoid significant semantic gaps between adjacent levels. In this process, the model learns a method to suppress inconsistency by assimilating conflicting spatial filtering information, preserving valuable information before combining it, automatically learning weight parameters. It progressively integrates semantic and detailed information directly between lower-level and higher-level features, avoiding information loss or degradation during multilevel transmission. This enhances scale feature invariance while incurring minimal computational overhead and ensuring simplicity of implementation
\subsubsection{Adaptive spatial fusion} 
AFPN utilizes ASFF (adaptively spatial feature fusion) \cite{liu2019learning} to assign different spatial weights to features at different levels of the multilevel feature fusion process in order to enhance the importance of the key levels and to mitigate the effect of contradictory information from different objects,as shown in Figure 5,
\(
x_{ij}^{n\rightarrow l}
\)
denotes the feature vector from level \(n\)to the location \(l\) of level \(\left( i,j \right)\),The resultant feature vector obtained by adaptive spatial fusion of multilayer features, denoted as  \(y_{ij}^{l}\), A linear combination of eigenvectors 
\(
x_{ij}^{1\rightarrow l},x_{ij}^{2\rightarrow l}
\) 
and  \(x_{ij}^{3\rightarrow l}\) defined as follows:
\begin{equation}
\
y_{ij}^{l}=\alpha _{ij}^{l}\cdot x_{ij}^{1\rightarrow l}+\beta _{ij}^{l}\cdot x_{ij}^{2\rightarrow l}+\gamma \cdot x_{ij}^{3\rightarrow l}\ 
\
\end{equation}
where \(y_{ij}^{l}\) r fers to the mapping of \(y^l\) in the output \(\left( i,j \right) \)  feature map between channels,\(\alpha _{ij}^{l},\beta _{ij}^{l}\)and \(\gamma _{ij}^{l}\) represent the spatial weights of the three hierarchical features at level  \(l\). These weights are obtained by adaptive learning of the network, where it is enforced that \(\alpha _{ij}^{l}+\beta _{ij}^{l}+\gamma _{ij}^{l}=\ 1\) and \(\alpha _{ij}^{l},\beta _{ij}^{l},\gamma _{ij}^{l}\in \left[ 0,1 \right] \), and defined that
\begin{equation}
\
\alpha _{ij}^{l}=\frac{e^{\lambda _{\alpha _{ij}}^{l}}}{e^{\lambda _{\alpha _{ij}}^{l}}+e^{\lambda _{\beta _{ij}}^{l}}+e^{\lambda _{\gamma _{ij}}^{l}}}
\
\end{equation}

Here \(\alpha _{ij}^{l},\beta _{ij}^{l}\) and \(\gamma _{ij}^{l}\) are defined by softmax function where \(\lambda _{\alpha _{ij}}^{l},\lambda _{\beta _{ij}}^{l}\) and \(\lambda _{\gamma _{ij}}^{l}\) are used as control parameters and a 1×1 convolutional layer is used to compute the weight scalar maps \(\lambda _{\alpha}^{l},\lambda _{\beta}^{l}\) and \(\lambda _{\gamma}^{l}\) from \(x^{1\rightarrow l},x^{2\rightarrow l}\) and \(x^{3\rightarrow l}\) respectively.

In addition, an adaptive spatial fusion module specific to a particular number of each stage is implemented in the AFPN. This approach allows the system to tailor the fusion strategy to the specifics of each fusion stage to more effectively deal with conflicting information between different layers and enhance the impact of key features. With this targeted fusion approach, the AFPN is able to better synthesize feature information from different levels in order to improve the accuracy and efficiency of target detection.
\begin{figure}[htb]
    \centering
    \includegraphics[width=1\textwidth, trim=1cm 4cm 2cm 2cm, clip]{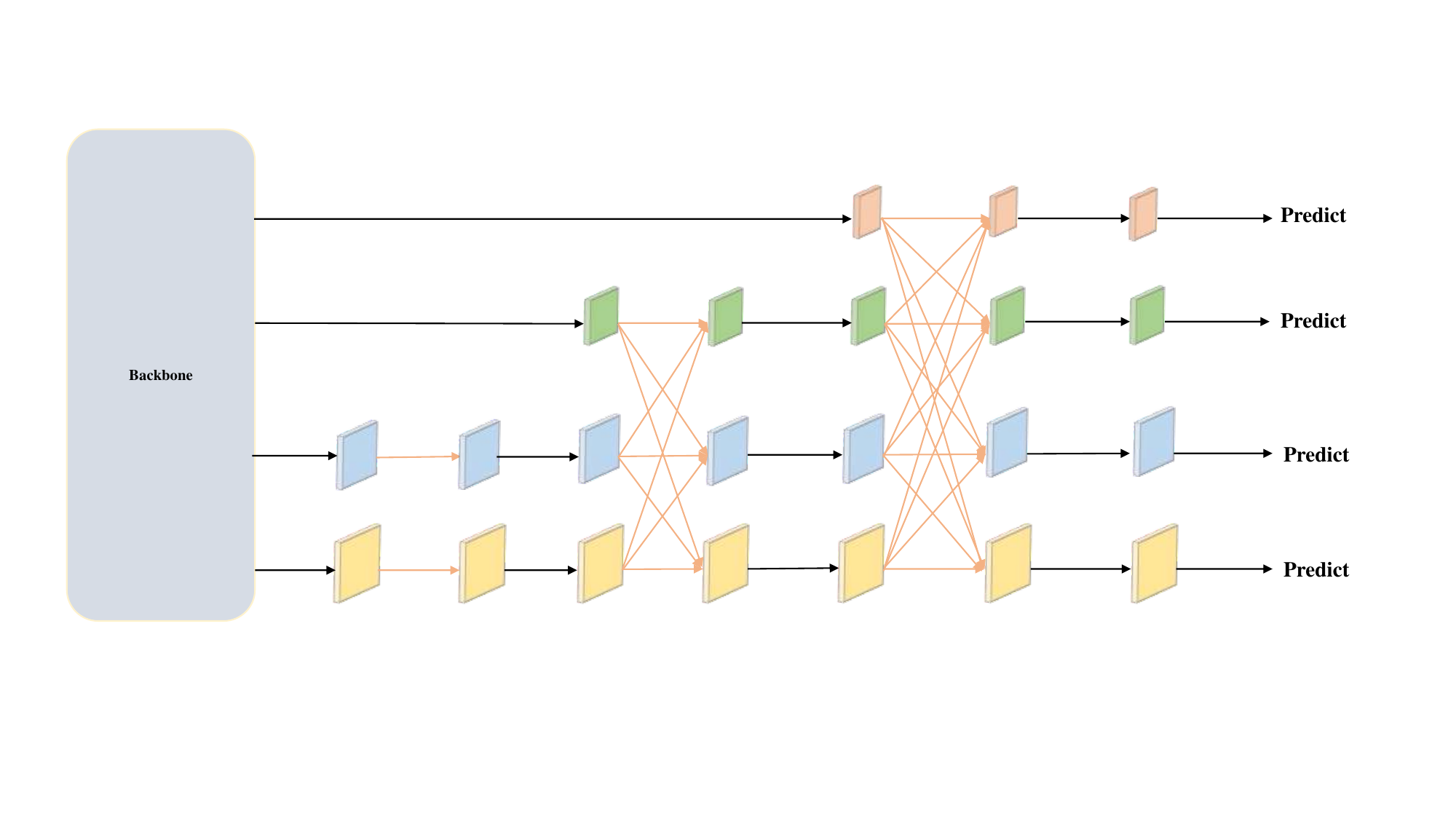}
    \caption{The architecture of the Asymptotic Feature Pyramid Network (AFPN).AFPN fuses two low-level features in the initial stage. Subsequent stages fuse high-level features, while the final stage adds top-level features during feature fusion. Black arrows indicate convolution and orange arrows indicate adaptive spatial fusion}
    \label{fig:AFPN在初始阶段融合两个底层特征。后续阶段融合高级特征，而最后阶段在特征融合过程中添加顶级特征。黑色箭头表示卷积，橙色箭头表示自适应空间融合}
\end{figure}
\begin{figure}[htb]
    \centering
    \includegraphics[width=1.05\textwidth, trim=3cm 1.5cm 3cm 3cm, clip]{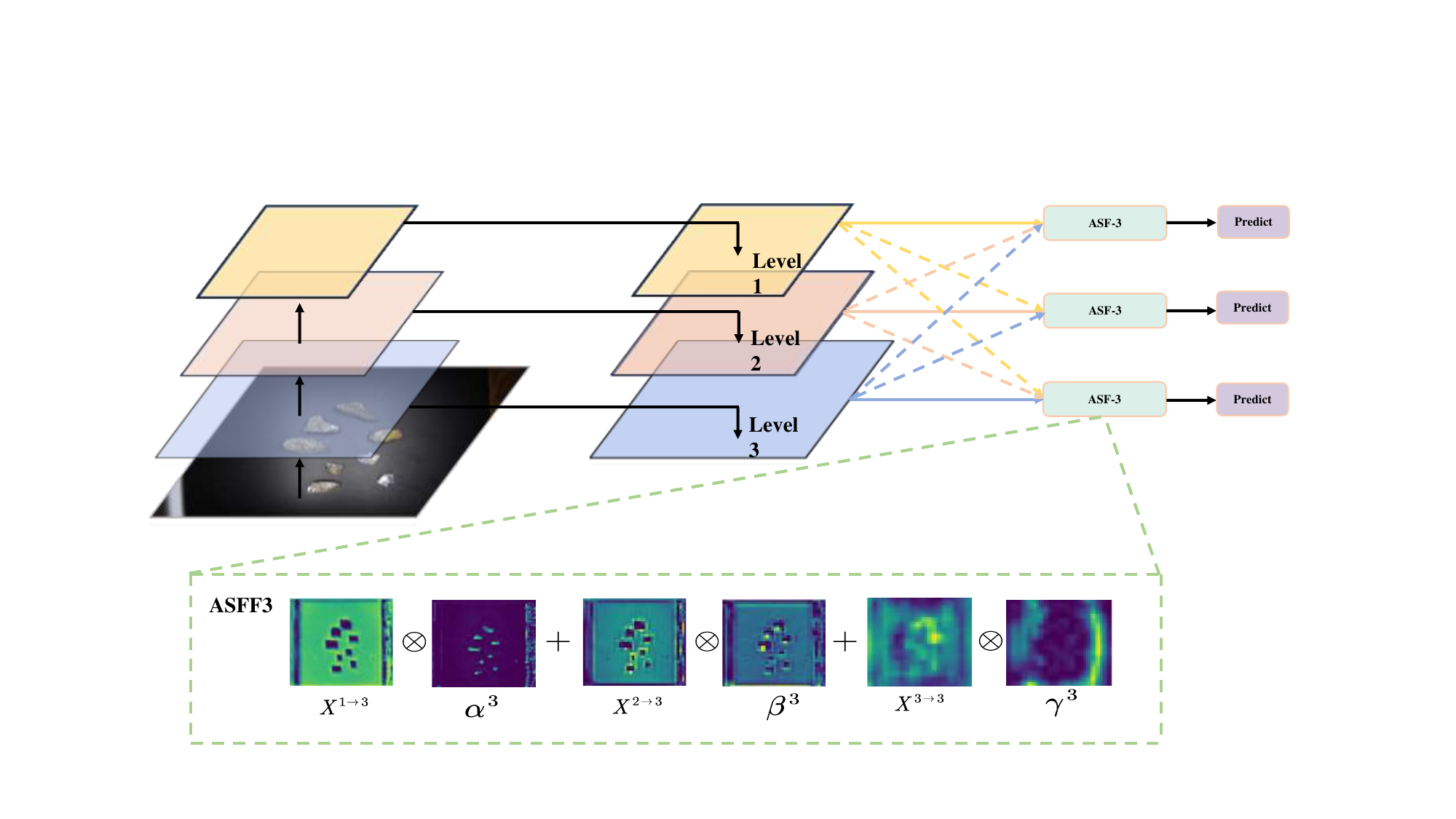}
    \caption{Description of the adaptive spatial feature fusion mechanism. The original up-sampling method is replaced with adaptive fusion, where for each layer the features of all other layers are adapted to the same shape by convolution and spatially fused based on a map of learned weights}
    \label{fig:自适应空间特征融合机制说明。将原先的上采样方法替换为自适应融合，对于每一层来说，其他所有其他层的特征都通过卷积被调整为相同的形状，并根据学习到的权重图进行空间融合}
\end{figure}
\subsection{ LOSS function}
The loss function of OreYOLO (Equation 4) comprises three components: bounding box regression loss, classification loss, and objectness loss.
\begin{equation}
\mathcal{L}_{\text{OreYOLO}}(t_p, t_{gt}) = \sum_{k=0}^{K \times K} \left[ \alpha_{k}^{\text{balance}} \alpha_{\text{box}} \sum_{i=0}^{S^2} \sum_{j=0}^B \mathbb{I}_{kij}^{\text{obj}} \mathcal{L}_{\text{MPDIoU}} + \alpha_{\text{obj}} \sum_{i=0}^{S^2} \sum_{j=0}^B \mathbb{I}_{kij}^{\text{obj}} \mathcal{L}_{\text{obj}} + \alpha_{\text{cls}} \sum_{i=0}^{S^2} \sum_{j=0}^B \mathbb{I}_{kij}^{\text{obj}} \mathcal{L}_{\text{cls}} \right]
\end{equation}

The network divides each feature map into several cells, and each cell outputs a vector of \(\left[ t_x,t_y,t_w,t_h,p_0\cdots \right] \) , where \(t_x,t_y\) are used to calculate the offsets between the centers of the prediction box and the corresponding anchor box, \(t_w,t_h\) is used to calculate the width and height of the prediction box, and \(p_0\) is the probability that the cell contains the target.\(K,S^2,B\) represents the output feature map, the number of cells into which the output feature map is divided, and the number of anchors on each cell, respectively. \(\alpha \) represents the weights of different components (such as the weights of localization loss, confidence loss, and classification loss), which are used to balance the weights of the output feature maps at each scale. The default value of \(\alpha _{k}^{balance} \) is corresponding to the output feature maps of 80 × 80, 40 × 40, and 20 × 20, respectively.

\begin{equation}
\
d_{1}^{2}=\left( x_{1}^{prd}-x_{1}^{gt} \right) ^2+\left( y_{1}^{prd}-y_{1}^{gt} \right) ^2
\
\end{equation}

\begin{equation}
\
d_{2}^{2}=\left( x_{2}^{prd}-x_{2}^{gt} \right) ^2+\left( y_{2}^{prd}-y_{2}^{gt} \right) ^2
\
\end{equation}

\begin{equation}
\
MPDIoU=IoU-\frac{d_{1}^{2}}{h^2+w^2}-\frac{d_{2}^{2}}{h^2+w^2}
\
\end{equation}

\begin{equation}
\
\mathcal{L}_{MPDIoU}=1-MPDIoU
\
\end{equation}

The bounding box regression loss adopts the MPDIoU \cite{siliang2023mpdiou} loss function (Equations 5-8), which encompasses all relevant factors considered in existing loss functions, namely overlap or non-overlap area, center point distance, as well as deviations in width and height, while simplifying the computation process. Here, \(d_{1}^{2}\) and \(d_{2}^{2}\) calculate the squared differences between the center points of the predicted box (\(pred\)) and the ground truth box (\(gt\)) in the \(x\) and \(y\) Ddirections, respectively, \(h\) and \(w\) are the heights and widths of the bounding boxes, respectively.

\begin{equation}
\
\sigma \left( x \right) =Sigmoid\left( x \right) =\frac{1}{1+e^{-x}}
\
\end{equation}

\begin{equation}
\
\mathcal{L}_{cls}=-\frac{1}{N}\sum_{i=1}^M{\left[ y_i\log \left( \hat{y}_i \right) +\left( 1-y_y \right) \log \left( 1-\hat{y}_i \right) \right]}
\
\end{equation}

where \(C\) is the total number of categories, \(y_i\) is the true label of the \(i\) category (1 if the target belongs to that category, 0 otherwise), and \(\hat{y}_i\) is the probability that the network predicts that the target belongs to the \(i\) category

\begin{equation}
\
\mathcal{L}_{obj}=-\frac{1}{M}\sum_{j=1}^M{\left[ c_j\log \left( \hat{c}_j \right) +\left( 1-c_j \right) \log \left( 1-\hat{c}_j \right) \right]}
\
\end{equation}

where,\(M\) is the total number of predicted bounding boxes,\(c_j\) is the true confidence of the \(j\) bounding box,\(\hat{c}_j\) is the confidence that the network predicts that the \(j\) bounding box contains the target.

\section{EXPERIMENTS}
\subsection{Image dataset preparation}
The acquisition of high-quality image data is crucial in the process of realizing the accurate classification task. In this study, a manual acquisition strategy was adopted to collect the required images, and the initial work encompassed the crushing of the construction site ores, which were cleaned, screened and de-dusted by a conveyor belt, which effectively minimized the negative impact of the ore surface dust on the image quality. Two ore types were included in the sample set: sulfide iron ore and gold ore. In the selection of image acquisition equipment, Hikvision industrial cameras with high resolution were chosen for this study.The number of ore samples in the acquisition process was varied, ranging from a single to ten, to truly reflect the complexity of the industrial site. On this basis, the research team meticulously maintained the randomness of the type, number and location of the samples to increase the richness and realism of the dataset. Overall, a total of 1,913 high-definition ore images were collected and accurately labeled with the Labelimg tool, providing a solid data foundation for deep learning model training.
\subsubsection{Data augmentation methods}
In the YOLOv5-based mineral recognition task, a broad and diverse training dataset is essential to enhance the generalization and accuracy of the model. Considering this need, this study adopts a series of image enhancement techniques in the data preprocessing stage to ensure the extensiveness and complexity of the dataset. Specifically, by randomly combining six strategies such as noise introduction, image rotation, cropping, panning, reflection, and brightness adjustment, this study realizes the augmentation and diversification of the original image data. Each original image was augmented four times in different combinations, which increased the final image data volume to 6090 images, thus providing a richer and more varied learning base for model training. Subsequently, these images are divided into training, validation and testing sets in a 7:2:1 ratio to support model learning and evaluation at various stages

\subsection{Model setting details}
The overall architecture of OreYOLO is shown in Fig. 6 and consists of five parts: the input layer (Input), the backbone network (Backbone), the spatial pooling pyramid of cross-domain partial connections (SPPFCSPC), the neck network (Neck) and the prediction layer (YOLO Head). The backbone network consists of three feature extraction blocks including the basic convolutional module (CBM), the cross-domain partial module incorporating the attention mechanism (CSP3\underline{~}EMA), and the basic cross-domain partially connected module(CSP3).Specifically, the CBM serves as the basic convolutional block of the network consisting of a convolutional layer and a Mish activation function, and we embedded the EMA in the CSPNet attention block, aiming at the network's ability to better fuse features of different channel dimensions when performing feature extraction. In addition, considering that horizontal feature fusion is also a preferred strategy to enhance the model's representational capability, we added the SPPFCSPC \cite{wang2023yolov7} module to OreYOLO to implement horizontal feature fusion, which performs convolutional operations on the model's output feature maps through spatial pooling pyramids, and then passes and fuses the features through the cross-domain partial connectivity as a way of increasing the model's receptive field and preserving the model's the original feature information of the output. The neck network employs a progressive feature pyramid structure, in which ASFF  weights features at different levels to effectively fuse them, as described in detail in Section 3.2. The prediction part uses three multi-scale feature maps to detect objects in the input image.

\begin{figure}[htb]
    \centering
    \includegraphics[width=1\textwidth, trim=3.5cm 3.5cm 3.5cm 3.5cm, clip]{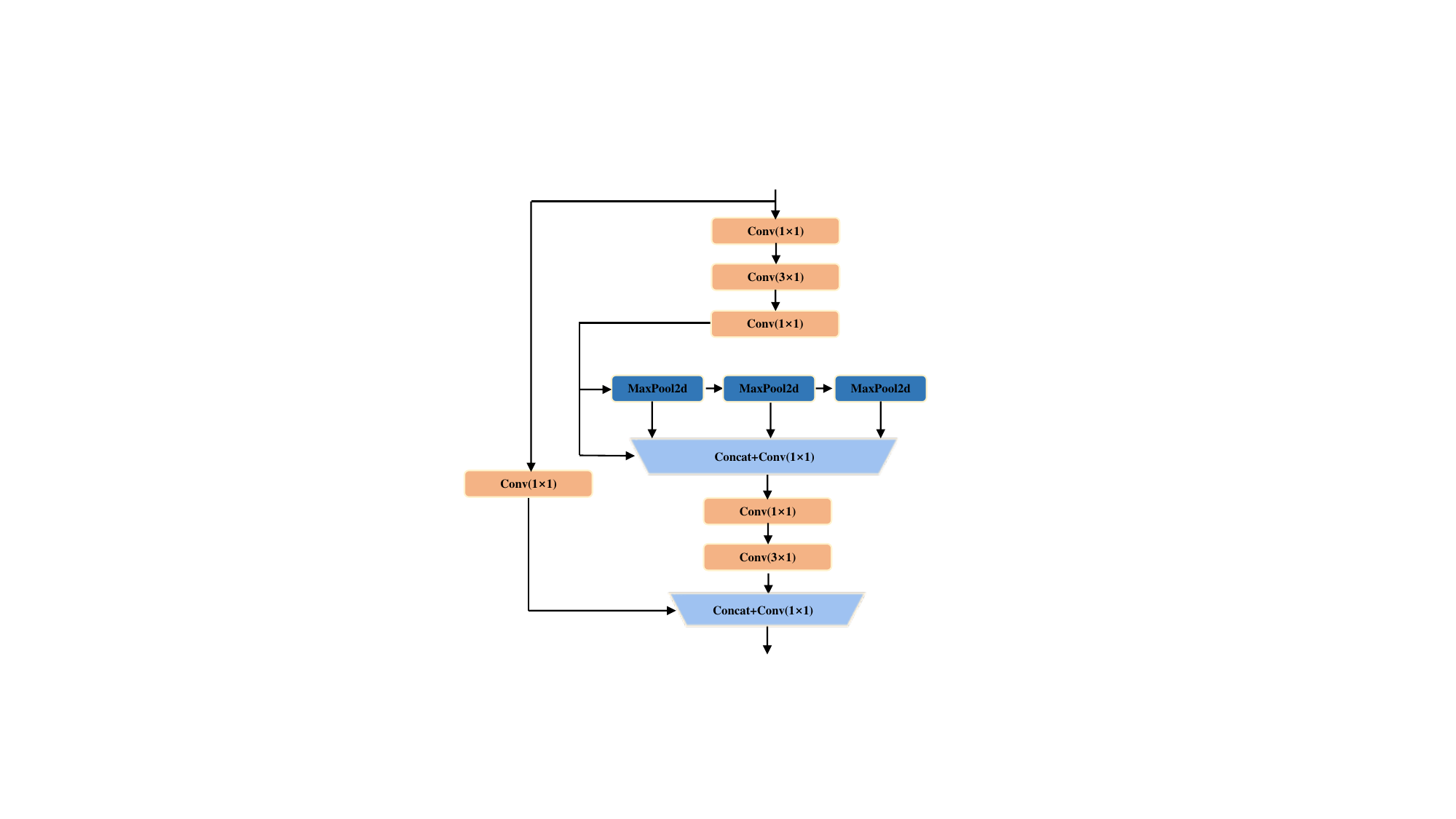}
    \caption{The structure of the SPPFCSPC modular,consisting of multiple convolutions, pooling operations and a residual path}
    \label{fig:SPPFCSPC模块结构，由多个卷积，池化操作和一个残差路径组成。}
\end{figure}
\begin{figure}[htb]
    \centering
    \includegraphics[width=1\textwidth, trim=1cm 1.5cm 1cm 1cm, clip]{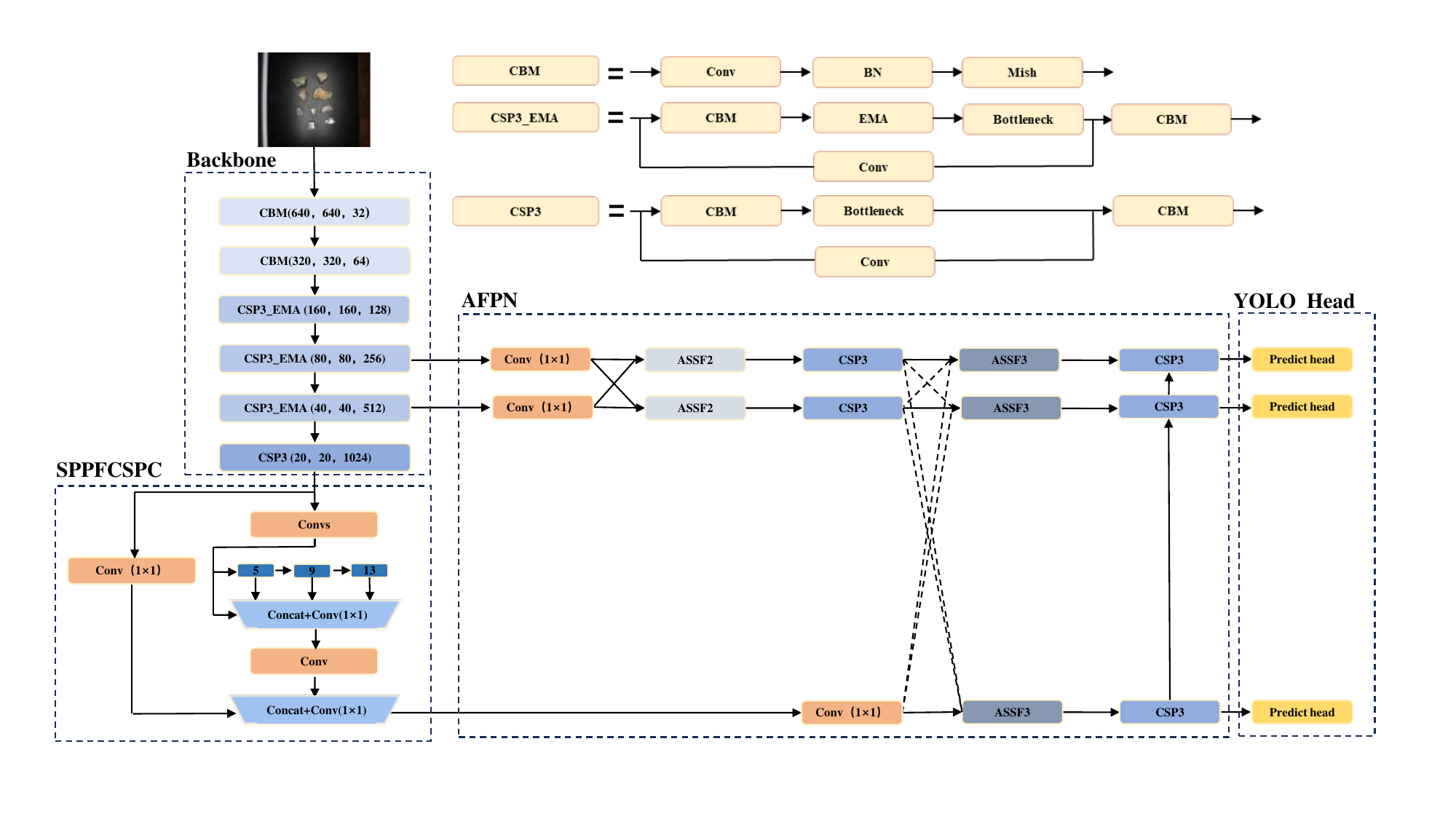}
    \caption{The architecture of OreYOLO,including five parts,input, backbone network CSP3\underline{~}EMA for feature extraction, SPPFCSPC for horizontal feature fusion, AFPN for vertical feature fusion, and YOLO Head for prediction}
    \label{fig:YOLOv5架构图，包括五部分，输入端，用于提取特征的骨干网络CSP3_EMA，用于水平特征融合的SPPFCSPC,用于垂直特征融合的AFPN以及用于预测的YOLO Head}
\end{figure}
\subsection{Result analysis}
In the result analysis, we build ten commonly used high-performance target detection networks for comparison experiments. The comparison models include one-level architectures such as Yolo V3 \cite{redmon2018yolov3}, Yolo V4 \cite{bochkovskiy2020yolov4}, YoloV5, EfficientDet \cite{howard2017mobilenets} and RetinaNet \cite{lin2017focal}. Two-tier architectures, such as Faster-RCNN \cite{ren2015faster} and unanchored architectures, such as CenterNet  \cite{duan2019centernet} and Yolo X \cite{ge2021yolox}, are modeled with the same details as presented in the corresponding original papers.

The experimental results of OreYOLO and other comparative networks in the gold and sulfide ore datasets are shown in Table 3. The mAP50-95 is used as one of the main criteria to measure the model performance in this experiment as a way to comprehensively evaluate the performance of the model under different IoU threshold conditions, so as to verify its robustness and accuracy in the target detection task. Specifically, OreYOLO can achieve 86.6\% mAP50-95 in the gold and sulfide ore datasets, which is 3.2\% higher than the third best Faster-RCNN (83.4\%), while OreYOLO still has a clear advantage in detection performance, e.g., compared with its peer series (YOLO X,YOLO v4,YOLO v3, etc.). OreYOLO also performs better than the best YOLO X (79.6\%), which is 7\% higher. In addition, OreYOLO differs from the best YOLO v5 (Small) by only 0.6\% on mAP50-95 while maintaining a low number of parameters and a low computational effort, which highlights the model's remarkable ability to discriminate between these two mineral types with high precision.

In addition, we also compared the model-related influencing factors such as computational complexity, model size and inference speed in detail in the comparison experiments, and the differences between OreYOLO and the other models in terms of model parameters (\#Param.), computational volumes (GFLOPs), etc. are given in Table 4, and the results of the experiments show that OreYOLO has relatively fewer parameters and lower GFLOPs, respectively, of 3.458M and 6.3G, proving its high performance in terms of model size and computational complexity, whereas for other single-stage, two-stage, and unanchored architectures, only EfficientDet-D0 (\#Param. of 3.83M and GFLOPs of 4.614G) can compete with OreYOLO.In addition to the inference speed, OreYOLO achieves the highest frame rate of 79.07, which is 5.77 higher than the best one-stage architecture (Yolo V5, with a frame rate of 73.26), and 18.19 higher than the no-anchor architecture (CenterNet, with a frame rate of 60.88).In addition, when combined with the above results on the detection accuracy, it is clear that OreYOLO strikes an excellent balance between detection accuracy and efficiency, while the small number of parameters and low computational complexity ensure its feasibility for deployment in most edge-complex environments.

\begin{table}[htb]
\caption{Experimental parameter settings}
\vspace{-5pt}
\centering
\setlength{\tabcolsep}{65pt}
\renewcommand{\arraystretch}{1.3} 
\begin{tabular}{ll}
\toprule
Parameter name   & Selected value \\
\midrule
depth\_multiple    & 0.20           \\
width\_multiple    & 0.25           \\
Input shape        & 640×640        \\
Epoch              & 100            \\
Optimizer          & AdamW          \\
learning rate      & 1e-3           \\
momentum           & 0.937          \\
Nms\_iou           & 0.45           \\
label-smoothing    & 0.005          \\
Confidence         & 0.25           \\
Mixup probability  & 0.5          \\
Mosaic probability & 0.5          \\
\bottomrule
\end{tabular}
\label{tab:实验参数表}
\end{table}

\begin{table}[htb]
\caption{Different performance of OreYOLO on gold and sulfide ore datasets)}
\vspace{-5pt}
\centering
\setlength{\tabcolsep}{20pt}
\renewcommand{\arraystretch}{1.3} 
\begin{tabular}{llllll}
\toprule
Type             &Precision    &Recall    &mAP50  &mAP75  &mAP50-95\\
\midrule
Au                & 99.3\%     & 99.3\%   &99.4\%  &99.4\%   &85.7\%\\
S                 & 99.2\%     & 99.2\%   &99.4\%  &99.1\%   &87.5\% \\

\bottomrule
\end{tabular}
\label{tab:金矿和硫铁矿参数表}
\end{table}

\begin{table}[htb]
\caption{comparison of the performance of OreYOLO with other classical target detection algorithms on the gold and sulfide ore datasets,including accuracy, recall, mAP50,mAP75,mAP50-95 (0.05 intervals)}
\vspace{-5pt}
\centering
\setlength{\tabcolsep}{20pt}
\renewcommand{\arraystretch}{1.3} 
\begin{tabular}{llllll}
\toprule
Series           &Type             &Edition     &mAP50      &mAP75     &mAP50-95\\
\midrule
                & Yolo X           &Tiny        & 98.2\%      &97.6\%    &79.6\%\\
                & Yolo V5          &Small       & 99.3\%      &99.1\%    &87.4\%\\
                & Yolo V4          &Darknet 53  & 95.8\%      &89.3\%    &68.5\%\\
One-stage       & Yolo V3          &Darknet 53  & 97.1\%      &88.4\%    &69.1\%\\
                &                  &D0          & 97.5\%      &95.6\%    &77.0\%\\
                & EfficientDet     &D1          & 96.0\%      &84.7\%    &67.6\%\\
                &                  &D2          & 96.3\%      &85.2\%    &68.0\%\\
                & RetinaNet        &Resnet 50   & 99.2\%      &98.0\%    &82.2\%\\
 \hline               
Two-stage       & Faster-RCNN      &Resnet 50   & 98.8\%      &97.6\%    &83.4\%\\
Anchor-free     & Yolo X           &Tiny        & 98.2\%      &97.6\%    &79.6\%\\
                & CenterNet        &Resnet 50   & 98.2\%      &95.8\%    &79.6\%\\
Proposed        &                  &CSP\_EMA    & 99.3\%      &99.2\%    &86.6\%\\                     
\bottomrule
\end{tabular}
\label{tab:对比实验1}
\end{table}
\begin{table}[htb]
\caption{Comparison between OreYOLO and other classical target detection algorithms on \#Param,GFLOPs,FPS,size is the size of the network input image size}
\vspace{-5pt}
\centering
\setlength{\tabcolsep}{16pt}
\renewcommand{\arraystretch}{1.3} 
\begin{tabular}{lllllll}
\toprule
Series           &Type            &Edition     &\#Param      &GFLOPs         & Size          &FPS\\
\midrule 
                & Yolo X           &Tiny        &  8.94M      &26.76G        &640          &47.68\\
                & Yolo V5          &Small       & 7.074M      &16.411G       &640          &73.26\\
                & Yolo V4          &Darknet 53  & 63.959M     &59.787G       &416          &17.06\\
One-stage       & Yolo V3          &Darknet 53  & 61.545M     &155.549G      &640          &38.06\\
                &                  &D0          & 3.838M      &4.614G        &512          &15.70\\
                & EfficientDet     &D1          & 6.558M      &11.219G       &650          &14.63\\
                &                  &D2          & 8.010M      &20.132G       &768          &12.41\\
                & RetinaNet        &Resnet 50   & 36.413M     &146.311G      &600          &36.57\\
\hline               
Two-stage       & Faster-RCNN      &Resnet 50   & 136.771M    &370.013G      &600          &21.44\\
Anchor-free     & Yolo X           &Tiny        &  8.94M      &26.76G        &640          &47.68\\
                & CenterNet        &Resnet 50   & 32.665M     &69.976G       &512          &60.88\\
Proposed        &                  &CSP\_EMA    & 3.458M      &6.3G          &640          &79.07\\                  
\bottomrule
\end{tabular}
\label{tab:对比实验2}
\end{table}
\subsection{Ablation experiment}
In the ablation experiments, we first compared in detail the effects of the EMA attention mechanism and the asymptotic feature pyramid network on OreYOLO in terms of detection performance, which were mainly evaluated in terms of mAP50-95 and the number of modeled parameters, as shown in Table 5. The addition of EMA and AFPN improves the mAP50-95 of OreYOLO by 1.3\% and 0.8\%, respectively, compared to the base network, while these improvements only increase the number of parameters by a small amount (0.039M and 0.138M) compared to the baseline network.

Table 6 compares the experimental results of OreYOLO with different attention mechanisms. We compare in detail the differences between CA, ECA, CBAM, and EMA attention mechanisms in terms of the number of parameters and accuracy. Specifically, CBAM and EMA improved the mAP50-95 of OreYOLO by 0.3\% and 0.8\%, respectively, and only slightly increased the number of parameters, whereas the performance of CA and ECA decreased slightly (by 0.1\% and 0.2\%, respectively). These results show that the EMA attention mechanism has better feasibility than other attention mechanisms in our network. In addition, we also compare the effects of different levels of feature fusion strategies on the performance of OreYOLO, which are detailed in Tables 5 and 7. From the experimental results in Table 5, it can be seen that SPPCSPC achieves a similar effect to AFPN in terms of improving the accuracy (mAP50-95 improves by 0.7\%, while AFPN improves by 0.8\%). Table 7 further compares the differences between SPPFCSPC and SPPF modules in terms of accuracy, number of parameters, and computational effort. The results show that although SPPFCSPC performs well in improving accuracy (1\% improvement in mAP50-95), it brings more number of parameters and computation compared to SPPF.

Taken together, the results of the ablation experiments show the important impact of the attention mechanism and feature fusion strategy on the model performance. These findings provide a useful reference for the optimization of lightweight target detection networks, helping to select appropriate improvement strategies to strike a balance between accuracy improvement and parameter efficiency.

\begin{table}[htb]
\caption{Influence of EMA,AFPNM,SPPFCSPC modules on model performance}
\vspace{-5pt}
\centering
\setlength{\tabcolsep}{19.5pt}
\renewcommand{\arraystretch}{1.3} 
\begin{tabular}{llllllll}
\toprule
&                 &EMA           &AFPN                 &SPPFCSPC      &mAP50-95     &\#Param\\
\midrule
&Base             &               &                    &               &84.8\%     &1.710M\\
&Base             &               &                    & \checkmark    &85.5\%     &3.317M\\
&Base             &               & \checkmark         &               &85.6\%     &1.848M\\
&Base             & \checkmark    &                    &               &86.1\%     &1.749M\\
&Base             & \checkmark    & \checkmark         & \checkmark    &86.6\%     &3.458M\\

\bottomrule
\end{tabular}
\label{tab:消融实验1}
\end{table}

\begin{table}[htb]
\caption{Influence of different attention mechanisms on model performance.}
\vspace{-5pt}
\centering
 \setlength{\tabcolsep}{7mm}{
\setlength{\tabcolsep}{14pt}
\renewcommand{\arraystretch}{1.3} 
\begin{tabular}{lllllll}
\toprule
               & CA           & ECA      & CBAM            & EMA                & mAP50-95      & \#Param \\
\midrule
 Base(AFPN+SPPFCSPC)     &            &               &            &              & 85.8\%       & 3.455M\\
 Base(AFPN+SPPFCSPC)     & \checkmark &               &            &              & 85.6\%       & 3.458M\\
 Base(AFPN+SPPFCSPC)     &           & \checkmark     &            &              & 85.7\%       & 3.455M\\ 
 Base(AFPN+SPPFCSPC)     &           &                & \checkmark &              & 86.1\%       & 3.456M\\
 Base(AFPN+SPPFCSPC)     &           &                &            & \checkmark   & 86.6\%      & 3.458M\\
\bottomrule
\end{tabular}
\label{tab:消融实验2}}
\end{table}
\begin{table}[htb]
\caption{Influence of SPPF,SPPFCSPC Modules on Model Performance.}
\vspace{-5pt}
\centering
\setlength{\tabcolsep}{17.5pt}
\renewcommand{\arraystretch}{1.3} 
\begin{tabular}{lllllll}
\toprule
                    & SPPF         & SPPFCSPC       & mAP50-95    &\#Param    &GFLOPs\\
\midrule
 Base(AFPN+EMA)     & \checkmark &                  & 85.6\%     & 1.851M     & 5.0G\\
 Base(AFPN+EMA)     &           & \checkmark        & 86.6\%     & 3.458M     & 6.3G\\

\bottomrule
\end{tabular}
\label{tab:消融实验3}
\end{table}


\section{Detection performance visualization}
The visualization results demonstrate the actual results of OreYOLO under the ore collection platform, as shown in Figure 9. In this scenario, OreYOLO is able to effectively differentiate between gold and sulfide ores, and the predicted bounding box precisely fits the contours of the ores, demonstrating its ability to accurately localize the ores. In addition, OreYOLO is able to accurately distinguish between difficult to distinguish ores (as shown by the yellow box in Figure 9) due to its excellent detection performance. The confidence parameter of the predicted object is also shown in the prediction box. From the visualization results, it can be observed that OreYOLO not only achieves accurate recognition in the face of difficult-to-distinguish ore blocks, but also has a high confidence level, fully demonstrating its effectiveness. Therefore, the above visualization results indicate that the attention mechanism, progressive feature fusion and horizontal feature fusion in OreYOLO together significantly enhance the feature extraction capability of the whole network.

\begin{figure}[htb]
    \centering
    \begin{minipage}[t]{0.48\textwidth}
    \centering
    \includegraphics[width=1.2\textwidth, trim=1cm 3cm 1cm 1cm, clip]{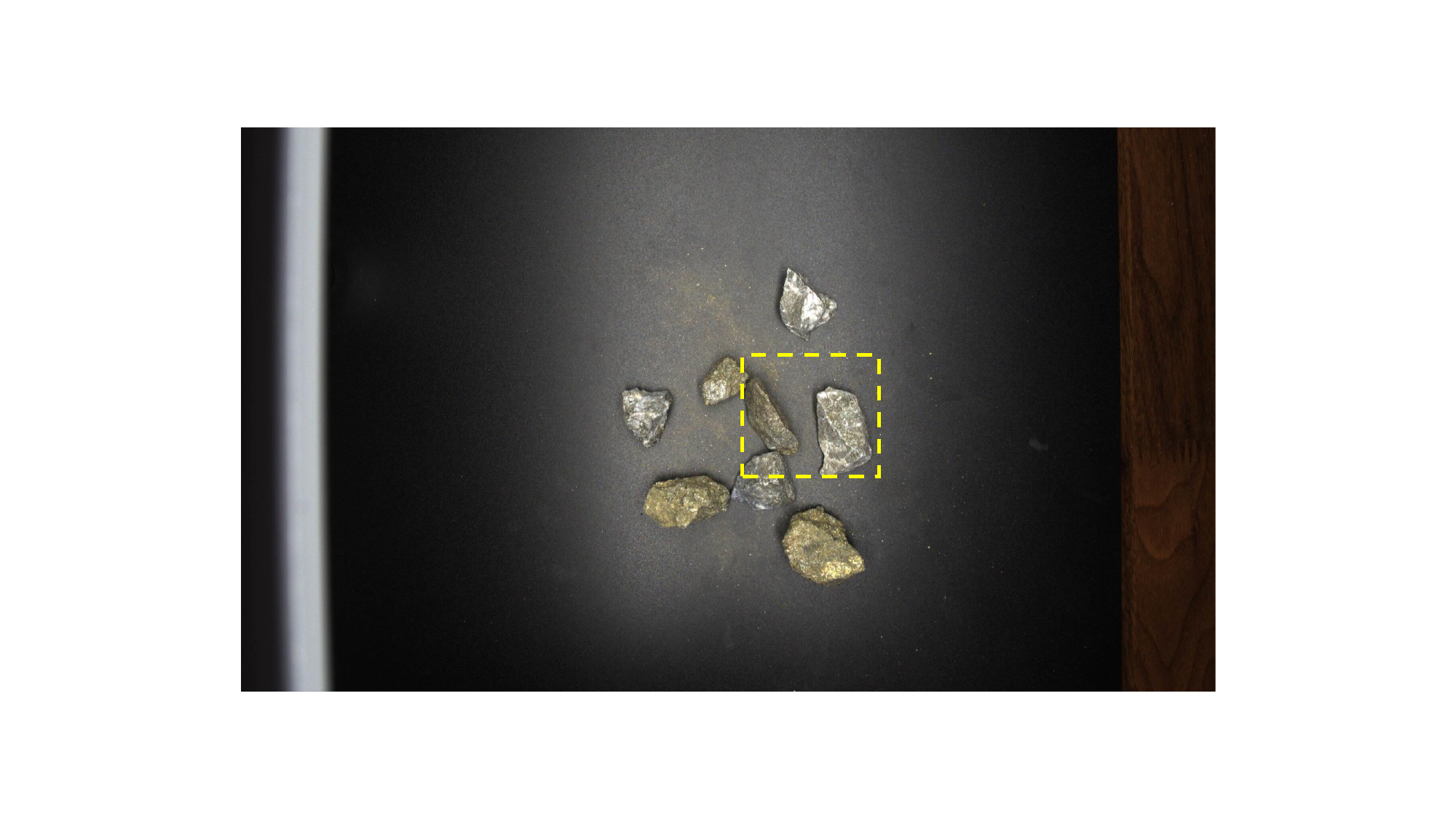}
   \end{minipage}
    \begin{minipage}[t]{0.48\textwidth}
    \centering
    \includegraphics[width=1.2\textwidth, trim=1cm 3cm 1cm 1cm, clip]{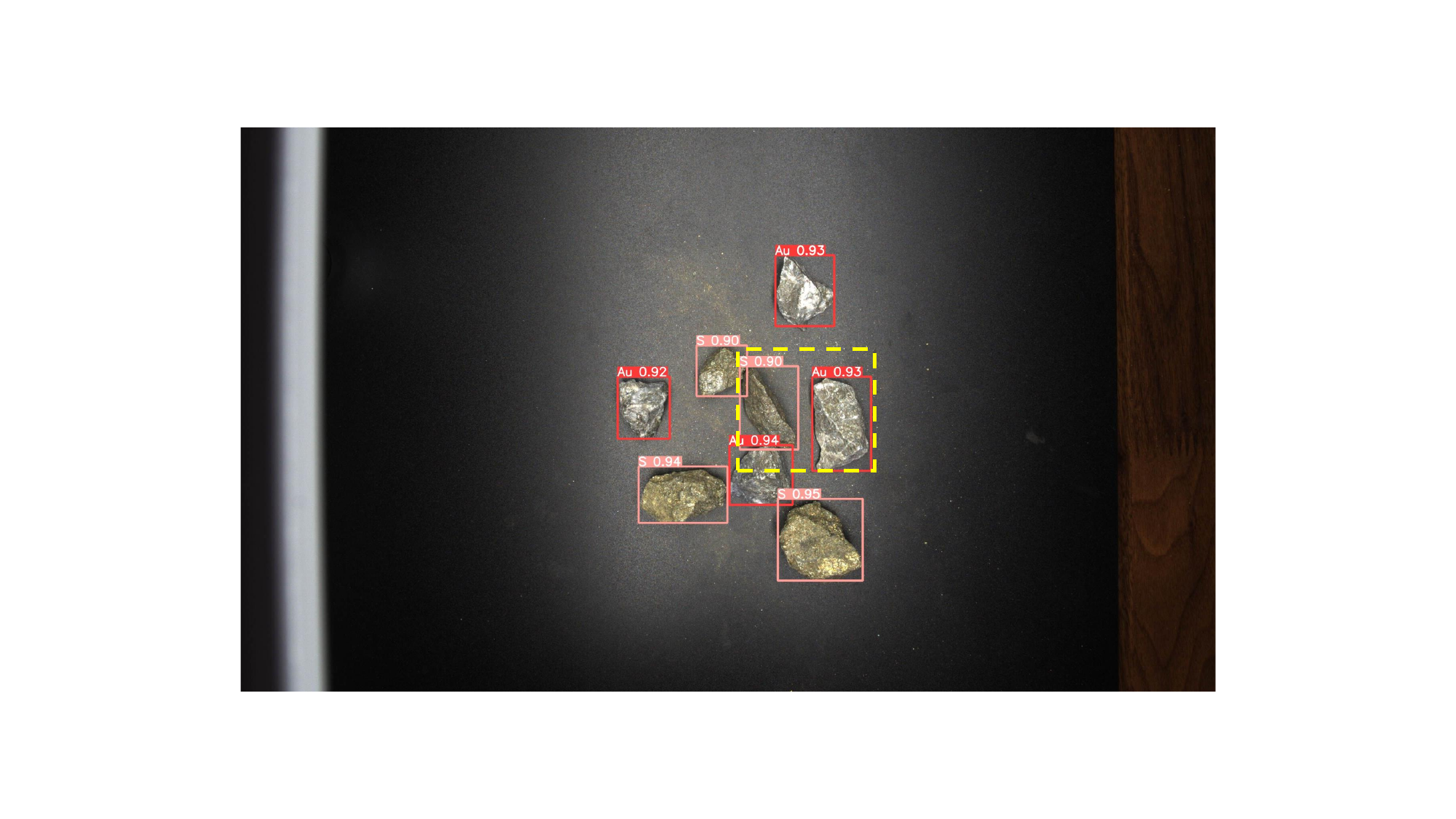}
   \end{minipage}
    \caption{Experimental data captured by an industrial camera, with the input image source on the left and the visualization results on the right}
    \label{fig:由工业相机采集的实验数据，左图为输入图像源，右图为可视化结果.}  
\end{figure}
\section{Conclusion}
While OreYOLO has demonstrated excellent performance on gold and sulphide ore data and can be successfully deployed to run in our facility environment, there are still some issues that need to be optimized.

First, deep learning-based ore sorting methods are very dependent on large amounts of clean data. Although we have performed extensive data augmentation to increase the diversity of our training samples, a significant amount of time and effort is still required to collect and prepare this data. This may involve collecting more samples in the field and ensuring the quality and diversity of the data to improve the accuracy and generalizability of the model.Second, the generalizability and robustness of the model needs to be continuously optimized and improved to ensure reliability and stability in real-world scenarios. This may require adjusting the model architecture, optimizing hyperparameters, and exploring more effective training strategies so that the model can better adapt to different operating conditions and ore types.Another challenge is model deployment and maintenance. Although OreYOLO performs well in a device environment, successfully deploying a deep learning model into a real device is not a simple task. Device resource constraints, the inference speed of the model, and its stability and reliability in real-world operation need to be considered. This may require optimizing the inference efficiency of the model, using lightweight models or model compression techniques to ensure efficient operation and stability of the model on the device.

Therefore, we need to continue our efforts to optimize the data collection and preparation process, as well as to continuously improve the training strategy and architecture of the models, as well as to enhance the stability and efficiency of the models during the deployment and operation phases.

\newpage
\bibliographystyle{unsrt}  
\bibliography{myread}  

\begin{thebibliography}{10}

\bibitem{su2018research}
Lingling Su, Xiangang Cao, Hongwei Ma, and Ying Li.
\newblock Research on coal gangue identification by using convolutional neural network.
\newblock In {\em 2018 2nd IEEE Advanced Information Management, Communicates, Electronic and Automation Control Conference (IMCEC)}, pages 810--814. IEEE, 2018.

\bibitem{liguan2020beneficiation}
Wang Liguan, Chen Sijia, Jia Mingtao, and T~Siyu.
\newblock Beneficiation method of wolframite image recognition based on deep learning.
\newblock {\em Chin. J. Nonferrous Met.}, 30(5):1192--1201, 2020.

\bibitem{zhang2019intelligent}
Ye~Zhang, Mingchao Li, Shuai Han, Qiubing Ren, and Jonathan Shi.
\newblock Intelligent identification for rock-mineral microscopic images using ensemble machine learning algorithms.
\newblock {\em Sensors}, 19(18):3914, 2019.

\bibitem{liu2021deep}
Yang Liu, Zelin Zhang, Xiang Liu, Wang Lei, and Xuhui Xia.
\newblock Deep learning based mineral image classification combined with visual attention mechanism.
\newblock {\em IEEE Access}, 9:98091--98109, 2021.

\bibitem{liu2023losn}
Yang Liu, Xueyi Wang, Zelin Zhang, and Fang Deng.
\newblock Losn: lightweight ore sorting networks for edge device environment.
\newblock {\em Engineering Applications of Artificial Intelligence}, 123:106191, 2023.

\bibitem{zhao2022machine}
Pengfei Zhao, Zhengjie Luo, Jiansu Li, Yujun Liu, and Baocheng Zhang.
\newblock Machine learning sorting method of bauxite based on se-enhanced network.
\newblock {\em Applied Sciences}, 12(14):7178, 2022.

\bibitem{zhang2022mineral}
Junyu Zhang, Qi~Gao, Hailin Luo, and Teng Long.
\newblock Mineral identification based on deep learning using image luminance equalization.
\newblock {\em Applied Sciences}, 12(14):7055, 2022.

\bibitem{huang2020weakly}
Zhijian Huang, Fangmin Li, Xidao Luan, and Zuowei Cai.
\newblock A weakly supervised method for mud detection in ores based on deep active learning.
\newblock {\em Mathematical problems in engineering}, 2020:1--10, 2020.

\bibitem{xiao2023mining}
Dong Xiao, Panpan Liu, Jichun Wang, Zhengmin Gu, and Hang Yu.
\newblock Mining belt foreign body detection method based on yolov4\_geca model.
\newblock {\em Scientific Reports}, 13(1):8881, 2023.

\bibitem{luo2022review}
Xianping Luo, Kunzhong He, Yan Zhang, Pengyu He, and Yongbing Zhang.
\newblock A review of intelligent ore sorting technology and equipment development.
\newblock {\em International Journal of Minerals, Metallurgy and Materials}, 29(9):1647--1655, 2022.

\bibitem{breiman1996bagging}
Leo Breiman.
\newblock Bagging predictors.
\newblock {\em Machine learning}, 24:123--140, 1996.

\bibitem{breiman2017classification}
Leo Breiman.
\newblock {\em Classification and regression trees}.
\newblock Routledge, 2017.

\bibitem{wang2007naive}
Qiong Wang, George~M Garrity, James~M Tiedje, and James~R Cole.
\newblock Naive bayesian classifier for rapid assignment of rrna sequences into the new bacterial taxonomy.
\newblock {\em Applied and environmental microbiology}, 73(16):5261--5267, 2007.

\bibitem{liu2019enhanced}
Chengzhao Liu, Mingchao Li, Ye~Zhang, Shuai Han, and Yueqin Zhu.
\newblock An enhanced rock mineral recognition method integrating a deep learning model and clustering algorithm.
\newblock {\em Minerals}, 9(9):516, 2019.

\bibitem{cover1967nearest}
Thomas Cover and Peter Hart.
\newblock Nearest neighbor pattern classification.
\newblock {\em IEEE transactions on information theory}, 13(1):21--27, 1967.

\bibitem{cortes1995support}
Corinna Cortes and Vladimir Vapnik.
\newblock Support-vector networks.
\newblock {\em Machine learning}, 20:273--297, 1995.

\bibitem{wang2020cspnet}
Chien-Yao Wang, Hong-Yuan~Mark Liao, Yueh-Hua Wu, Ping-Yang Chen, Jun-Wei Hsieh, and I-Hau Yeh.
\newblock Cspnet: A new backbone that can enhance learning capability of cnn.
\newblock In {\em Proceedings of the IEEE/CVF conference on computer vision and pattern recognition workshops}, pages 390--391, 2020.

\bibitem{lin2017feature}
Tsung-Yi Lin, Piotr Doll{\'a}r, Ross Girshick, Kaiming He, Bharath Hariharan, and Serge Belongie.
\newblock Feature pyramid networks for object detection.
\newblock In {\em Proceedings of the IEEE conference on computer vision and pattern recognition}, pages 2117--2125, 2017.

\bibitem{liu2018path}
Shu Liu, Lu~Qi, Haifang Qin, Jianping Shi, and Jiaya Jia.
\newblock Path aggregation network for instance segmentation.
\newblock In {\em Proceedings of the IEEE conference on computer vision and pattern recognition}, pages 8759--8768, 2018.

\bibitem{ouyang2023efficient}
Daliang Ouyang, Su~He, Guozhong Zhang, Mingzhu Luo, Huaiyong Guo, Jian Zhan, and Zhijie Huang.
\newblock Efficient multi-scale attention module with cross-spatial learning.
\newblock In {\em ICASSP 2023-2023 IEEE International Conference on Acoustics, Speech and Signal Processing (ICASSP)}, pages 1--5. IEEE, 2023.

\bibitem{hou2021coordinate}
Qibin Hou, Daquan Zhou, and Jiashi Feng.
\newblock Coordinate attention for efficient mobile network design.
\newblock In {\em Proceedings of the IEEE/CVF conference on computer vision and pattern recognition}, pages 13713--13722, 2021.

\bibitem{yang2023afpn}
Guoyu Yang, Jie Lei, Zhikuan Zhu, Siyu Cheng, Zunlei Feng, and Ronghua Liang.
\newblock Afpn: asymptotic feature pyramid network for object detection.
\newblock In {\em 2023 IEEE International Conference on Systems, Man, and Cybernetics (SMC)}, pages 2184--2189. IEEE, 2023.

\bibitem{liu2019learning}
Songtao Liu, Di~Huang, and Yunhong Wang.
\newblock Learning spatial fusion for single-shot object detection.
\newblock {\em arXiv preprint arXiv:1911.09516}, 2019.

\bibitem{siliang2023mpdiou}
Ma~Siliang and Xu~Yong.
\newblock Mpdiou: a loss for efficient and accurate bounding box regression.
\newblock {\em arXiv preprint arXiv:2307.07662}, 2023.

\bibitem{wang2023yolov7}
Chien-Yao Wang, Alexey Bochkovskiy, and Hong-Yuan~Mark Liao.
\newblock Yolov7: Trainable bag-of-freebies sets new state-of-the-art for real-time object detectors.
\newblock In {\em Proceedings of the IEEE/CVF conference on computer vision and pattern recognition}, pages 7464--7475, 2023.

\bibitem{redmon2018yolov3}
Joseph Redmon and Ali Farhadi.
\newblock Yolov3: An incremental improvement.
\newblock {\em arXiv preprint arXiv:1804.02767}, 2018.

\bibitem{bochkovskiy2020yolov4}
Alexey Bochkovskiy, Chien-Yao Wang, and Hong-Yuan~Mark Liao.
\newblock Yolov4: Optimal speed and accuracy of object detection.
\newblock {\em arXiv preprint arXiv:2004.10934}, 2020.

\bibitem{howard2017mobilenets}
Andrew~G Howard, Menglong Zhu, Bo~Chen, Dmitry Kalenichenko, Weijun Wang, Tobias Weyand, Marco Andreetto, and Hartwig Adam.
\newblock Mobilenets: Efficient convolutional neural networks for mobile vision applications.
\newblock {\em arXiv preprint arXiv:1704.04861}, 2017.

\bibitem{lin2017focal}
Tsung-Yi Lin, Priya Goyal, Ross Girshick, Kaiming He, and Piotr Doll{\'a}r.
\newblock Focal loss for dense object detection.
\newblock In {\em Proceedings of the IEEE international conference on computer vision}, pages 2980--2988, 2017.

\bibitem{ren2015faster}
Shaoqing Ren, Kaiming He, Ross Girshick, and Jian Sun.
\newblock Faster r-cnn: Towards real-time object detection with region proposal networks.
\newblock {\em Advances in neural information processing systems}, 28, 2015.

\bibitem{duan2019centernet}
Kaiwen Duan, Song Bai, Lingxi Xie, Honggang Qi, Qingming Huang, and Qi~Tian.
\newblock Centernet: Keypoint triplets for object detection.
\newblock In {\em Proceedings of the IEEE/CVF international conference on computer vision}, pages 6569--6578, 2019.

\bibitem{ge2021yolox}
Zheng Ge, Songtao Liu, Feng Wang, Zeming Li, and Jian Sun.
\newblock Yolox: Exceeding yolo series in 2021.
\newblock {\em arXiv preprint arXiv:2107.08430}, 2021.

\end{thebibliography}

\end{document}